\documentclass[10pt,journal]{IEEEtran}

\usepackage{amsmath,graphicx}
\usepackage{bm,amsmath,amssymb,amsfonts,epsfig,amsthm,color,tikz}
\usepackage{cite}
\usepackage{soul}
\usepackage{booktabs}
\usepackage{algorithm,algpseudocode,setspace}
\usepackage[algo2e]{algorithm2e}

\title{Learning Optimal Solutions for \\Extremely Fast AC Optimal Power Flow}

 \author{Ahmed S. Zamzam$^1$\thanks{$^1$National Renewable Energy Laboratory, Golden, CO, USA. Email: ahmed.zamzam@nrel.gov.} and Kyri Baker$^2$ \thanks{$^2$University of Colorado Boulder, Boulder, CO, USA. Email: kyri.baker@colorado.edu.}}

\begin{document}
 
\maketitle

\begin{abstract}
In this paper, we develop an online method that leverages machine learning to obtain feasible solutions to the AC optimal power flow (OPF) problem with negligible optimality gaps on extremely fast timescales (e.g., milliseconds), bypassing solving an AC OPF altogether. This is motivated by the fact that as the power grid experiences increasing amounts of renewable power generation, controllable loads, and other inverter-interfaced devices, faster system dynamics and quicker fluctuations in the power supply are likely to occur. Currently, grid operators typically solve AC OPF every 15 minutes to determine economic generator settings while ensuring grid constraints are satisfied. Due to the computational challenges with solving this nonconvex problem, many efforts have focused on linearizing or approximating the problem in order to solve the AC OPF on faster timescales. However, many of these approximations can be fairly poor representations of the actual system state and still require solving an optimization problem, which can be time consuming for large networks. In this work, we leverage historical data to learn a mapping between the system loading and optimal generation values, enabling us to find near-optimal and feasible AC OPF solutions on extremely fast timescales without actually solving an optimization problem.
\end{abstract}

\section{Introduction}
AC optimal power flow (OPF) problems are solved by grid operators in order to achieve the most economic generation dispatch to meet  network power demands while adhering to the physical constraints of the network. With the increasing integration of fluctuating renewable energy sources, system operators are required to perform more frequent adjustments of the generators set-points. Currently, real-time adjustments are realized using the automatic generation control (AGC) which may lead to sub-optimal operational points due to the use of restrictive affine control policies. Various linearizations or approximations of the AC OPF problem have been developed throughout the years to address the computational burden of solving the AC OPF in real time, but in many cases cannot guarantee feasibility of the approximate solution, and still require solving an optimization problem, which can be prohibitive as the network size grows. 

While not an exhaustive list, there are three general techniques that have been used in the literature to pursue OPF solutions on fast timescales \cite{Molzahn_list}: 1) Leveraging linearizations, approximations, or convexifications of the AC power flow equations (e.g. \cite{Lavaei10, swaroop2015linear, bolognani2015linear, linModels}); 2) Distributed techniques which allocate the computation of the AC OPF problem across devices, agents, or distributed computational platforms (e.g. \cite{Erseghe14, Zheng_Dist_16, BakerMPC}); and 3) So-called ``real-time" or ``online" optimization techniques (e.g., \cite{Tang17, opfPursuit, Hauswirth17}). Combinations of these three, i.e., distributed online approaches which utilize linearizations of the power flow equations, are also used \cite{opfPursuit, Bernstein_online_19}.

Very recently, machine learning (ML) approaches have been leveraged for solving difficult optimization tasks~\cite{Baker_jointCC, gunduz2019machine, wang2018review, Learning_ACOPF, Zamzam-19physics, yang2019real}. The data-driven nature of ML approaches is very pertinent as large quantities of measurements are being generated in modern systems but are not yet being fully utilized \cite{Data18}. In the area of power systems, ML approaches have been proposed to recover the power systems state \cite{Zamzam19}, to enhance solving optimal power flow problems (i.e., for learning a good starting point for AC OPF: \cite{Baker_learning}, for learning the active set of the DC OPF problem: \cite{Misra19, Deka19}) and for directly bypassing the use of solvers or iterative methods to learn solutions in an online fashion (i.e. leveraging ML to obtain the solution of mixed-integer quadratic programs: \cite{Bertsimas_learn_19}, and for learning a feasible and optimal DC OPF solution: \cite{DeepOPF}). 

In \cite{Deka19}, neural networks are used to learn a mapping from uncertainty realizations to the active set of a DC OPF problem as an intermediate step towards learning the optimal solution. Once the active set is determined, the optimal solution to the original problem can be recovered by solving a linear system of equations. This  work illustrated the benefits of using data to exploit the structure of the OPF problem and solve DC OPF on timescales appropriate for corrective control. However, showing the benefit of this particular framework for AC OPF problems, which are very difficult to solve in real time, has not yet been demonstrated. In addition, if the correct active set is not identified during the classification, the resulting solution of the DC OPF will be incorrect, and additional resources must be used for identifying the correct active set.

The work in \cite{DeepOPF} demonstrated that it is possible to learn the solution of the DC OPF problem directly and preserve feasibility. Encouraged by the extremely fast solution times for DC OPF with feasible solutions and negligible optimality losses when using a neural network in \cite{DeepOPF}, we explored the potential and limitations of learning a mapping from the system loading to optimal generation setpoints and voltages in AC OPF. Towards this goal, we developed a framework for using learning to obtain solutions to AC OPF problems that lends itself to real-time applications without using the DC approximation or other linearizations of the problem. 

Our approach thus offers the following benefits over traditional approaches to solve the the AC OPF problem and over learning approaches to solve the DC OPF problem:

\begin{itemize}
    \item There is no need to directly tackle the AC OPF optimization problem.
    \item No approximations, linearizations, or convexifications of the power flow equations are used.
    \item No distributed techniques or computationally intensive platforms are needed; the neural network maps system loading onto optimal generation values and voltage magnitudes in roughly 1 millisecond on a laptop computer (in the considered networks).
    \item Recovery of the full AC OPF solution is ensured through a fast iterative procedure, resulting in a final optimal and feasible solution.
\end{itemize}

Recently, the authors of~\cite{Learning_ACOPF} proposed a machine learning model that targets approximating the AC OPF solutions directly. The AC OPF problem is treated as a regression task where the generator set-points are estimated from the grid demand profile. Fixing the active power injection at all generators, including the slack bus, drastically affects the feasibility of the solution. Hence, the reported feasibility percentage is almost $50\%$ for a maximum of $10\%$ change in the base load profile in a small network. In addition, the design of the learning approach in~\cite{Learning_ACOPF} does not allow for a post-processing step where a feasible solution is obtained by refining the infeasible solution. In addition, the design of the test scenarios with only $10\%$ deviation without considering correlation makes the representativeness of the study questionable.

\begin{figure}
    \centering
    \includegraphics[width = 3.2in]{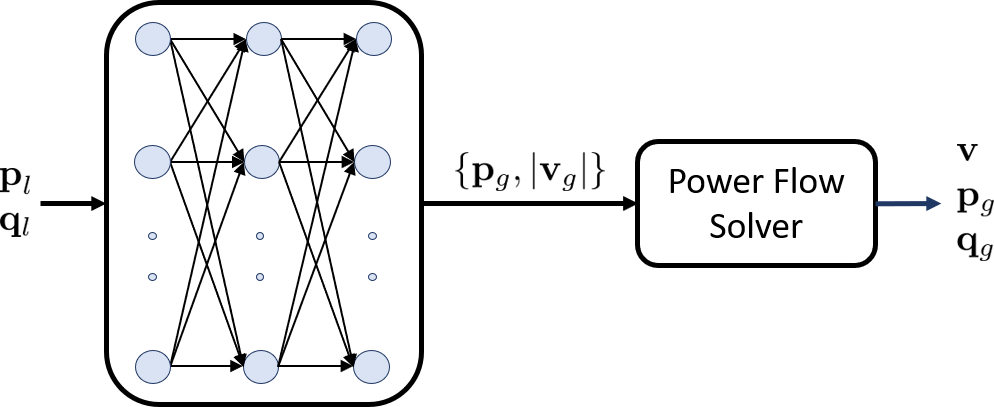}
    \caption{The flow of the proposed learning approach.}
    \label{fig:approach}
\end{figure}

The emergence of learning techniques in general can be attributed to the availability of increased computing power and an abundance of data collection and storage. Within the power systems realm, a large quantity of data is generated by grid operators by repeatedly solving AC OPF problems throughout the day. This data is utilized for other tasks, and often is not stored for long periods of time, as grid operators struggle to see additional use cases for the data \cite{Data18}. This paper demonstrates an additional purpose for this data by using it to train a neural network (offline) which can produce optimal solutions to the AC OPF problem (online). To achieve this, we learning a mapping from the network loads to the optimal generator set-points, where a novel parameterization of the AC OPF solutions is used to enforce generation limits. In addition, strictly feasible solutions are used to train the learning model, and the output of the model is constrained to adhere to active power generation and voltage magnitude constraints. In order to circumvent solutions that violate the system requirements, the solution feasibility is ensured by solving a set of power flow equations. Fig.~\ref{fig:approach} shows the architecture of the proposed approach in the operation phases. By using this framework, we can determine optimal generation set-points on timescales that are appropriate for balancing fast fluctuations in renewable generation and load.

The remainder of this paper is organized as follows. In Section II, the AC OPF problem formulation is presented. Then, the propoed learning approach is outlined in Section III, and the simulation results are presented in Section IV. Section V conculdes the paper and presents several directions for future research.


\section{Problem Formulation}\label{sec:ProbForm}
In this section, we will describe the AC OPF problem and our data-driven approach to reformulate the problem appropriately. First, consider a power network with $ N $ buses collected in set $ \mathcal{N} $, and set $ \mathcal{G}$ collects the set of nodes with generators, and set $\mathcal{L}$ collects the set of load buses. 
The nodal power injection at bus $ n  \in \mathcal{N}$ is denoted by $ p_{n} + j~q_{n}$. Also, let $p_{g,n}+j~q_{g,n}$, and $p_{l,n}+j~q_{l,n}$ denote the  power generated at bus $n\in\mathcal{G}$ and the load demand and bus $n\in\mathcal{L}$, respectively. Note that the sets $\mathcal{G}$ and $\mathcal{L}$ are both subsets of $\mathcal{N}$ but need not to be disjoint. The voltage at bus $n$ is denoted by $v_n$, and define ${\bf v} := [v_n]_{n\in\mathcal{N}}$, ${\bf p} := [p_n]_{n\in\mathcal{N}}$, and ${\bf q} := [q_n]_{n\in\mathcal{N}}$ which collect the voltage phasors, active power injection, and reactive power injection at all buses, respectively. Therefore, the AC OPF problem can be formulated as
\begin{subequations}\label{eq:OPF}
\begin{align}
    \underset{{\bf v}, \{p_{g,n}, q_{g,n}\}_{n\in\mathcal{G}}}{\text{minimize}} &\quad c({\bf p}_g)\ =\ \sum_{n\in\mathcal{G}} c_n(p_{g,n})\label{eq:cost}\\
    \text{subject to} & \qquad \underline{p}_{g,n} \leq p_{g,n} \leq \overline{p}_{g,n}\label{eq:gen-const}\\
     & \qquad \underline{q}_{g,n} \leq q_{g,n} \leq \overline{q}_{g,n}\label{eq:gen-const2}\\
     &\qquad \underline{| v |} \leq |{\bf v}| \leq \overline{| v |}\label{eq:vol-const}\\
     &\qquad {\bf h}({\bf v}, {\bf p}, {\bf q}) = {\bf 0}\label{eq:pfe}
\end{align}
\end{subequations}
where $ \underline{p}_{g,n}$ ($ \underline{q}_{g,n}$)  and $ \overline{p}_{g,n} $ ($ \overline{q}_{g,n} $)  denote the minimum and maximum active (reactive) power injection from the generator installed at bus $ n \in \mathcal{G}$, and $\underline{| v |}$ and $\overline{| v |} $ represent the lower and upper limits on the voltage magnitudes. The vector collecting the active power injection from all generators is denoted by ${\bf p}_g$, and the cost of generating power from the generator installed at bus $n$ is denoted by $c_n(p_{g,n})$. The \emph{nonlinear} power flow equations are collected in the equality constraint~\eqref{eq:pfe}.

Given the active and reactive power demand at the load buses, the active power generated by the generators, and the voltage magnitudes at the generator buses, the vector $ {\bf v}$ can be recovered by solving the power flow equations~\cite{Araposthatis-1981}. Utilizing this property, we treat the AC OPF problem as an operator with inputs representing the load profile while the generators' active power injections and voltage magnitudes are the output of this operator. In other words, we treat the OPF as an operator~\cite{zhou2019optimal} with inputs comprising $p_{l,n}$ and $q_{l,n}$ for all $n\in \mathcal{L}$ and with output representing $|v_{n}|$ for buses $n \in \mathcal{G}$ and $p_{g,n} $ for nodes $n\in \mathcal{G}\slash \{n_0\}$, where $n_0$ is the index of the reference bus. Therefore, the goal of the proposed data-driven approach is learn the underlying OPF mapping between the load demand at all load buses and the generators' active power and voltage magnitude set-points.

Any active power generation $p_{g,n}$ that satisfies~\eqref{eq:gen-const} can be written as 
\begin{align*}
    p_{g,n} = \underline{p}_{g,n}  + \alpha_n \bigg( {\overline{p}_{g,n} - \underline{p}_{g,n}}\bigg),
\end{align*}
where $0 \leq \alpha_n \leq 1$. Similarly, the voltage magnitude at bus $n\in\mathcal{G}$ can be parameterized by $\beta_n\in [0, 1]$ which is given by
\begin{align*}
    |v_n| =  {\underline{|v|} }+ \beta_n \bigg({\overline{|v|} - \underline{|v|}}\bigg).
\end{align*}
Define the sets $\mathcal{A} := \big\{[\alpha_n]_{n\in\mathcal{G}\slash \{n_0\}}\ | 0 \leq \alpha_n \leq 1 \big\}$ and $\mathcal{B} := \big\{[\beta_n]_{n\in\mathcal{G}}\ | 0 \leq \beta_n \leq 1 \big\}$.  Therefore, the AC OPF operator can be written as 
\begin{align}
\boldsymbol{\Omega} : \mathcal{P}_l \times \mathcal{Q}_l \rightarrow \mathcal{A} \times \mathcal{B},    
\end{align}
where $\mathcal{P}_l$ and $\mathcal{Q}_l$ represent the sets of active and reactive load demand at all buses $n\in \mathcal{L}$, respectively. 
Paremterizing the solution of the AC OPF problem using the parameters $\alpha_n$ and $\beta_n $ will prove useful when approximating the mapping $\Omega$ using artificial neural networks.  

\section{Learning feasible AC OPF Solutions}

Artificial neural networks (hereafter, NN) are well-suited for modeling high-dimensional, complex, non-linear relationships between input features and output variables. They are capable of considering massive amounts of training data (here, solutions to the AC OPF problem) and uncovering natural relationships in the data to perform classification or regression tasks extremely quickly. It is for this reason that they have been used previously for predicting OPF solutions \cite{DeepOPF, Learning_ACOPF} and the active set of OPF problems \cite{Deka19, Learning_ACOPF}. In this section, we discuss the generation of training data for the NN, training of the NN, and how the overall feasible solution to the AC OPF problem is obtained.


\subsection{Approximating the mapping between loads and the optimal solution}

In order to approximate the mapping $\boldsymbol{\Omega}$, we utilize a deep neural network model which leverages sigmoid activation functions. For finite-width deep neural networks, smooth functions can be approximated up to a specified accuracy which depends on the smoothness parameters of the function as well as the neural network size~\cite{grohs2019deep, Zamzam19}. That is, if a neural network $ g(\cdot ;\theta) $ is optimally trained with samples of $f(\cdot)$ on a closed set, then the distance between the output of the neural network $\hat{\bf y} = g(\cdot ; \theta)$ and ground-truth mapping ${\bf y} = f(\cdot)$ can be bounded as
\begin{align*}
    \| \hat{\bf y} - {\bf y} \|_\infty \leq \epsilon,
\end{align*}
where $\epsilon$ is function of the smoothness parameters of $f(\cdot)$ and the size of the neural network $g(\cdot;\theta)$. 
As the solution of AC OPF is often on the boundary of the feasibility set~\cite{molzahn2015semidefinite}, we aim at ensuring that the neural network outputs are in the interior of the AC OPF feasibility set. 

In order to avoid these cases where the voltage magnitude is exactly at the maximum or the lower voltage limits, we generate the training by solving the following restricted AC OPF problem (R-ACOPF).
\begin{subequations}\label{eq:ROPF}
\begin{align}
    \underset{{\bf v}, \{p_{g,n}, q_{g,n}\}_{n\in\mathcal{G}}}{\text{minimize}} \quad\sum_{n\in\mathcal{G}} &c_n(p_{g,n})\label{eq:ROPF-cost}\\
    \text{subject to}  \qquad \underline{p}_{g,n} &\leq p_{g,n} \leq \overline{p}_{g,n}\label{eq:ROPF-gen-const}\\
      \qquad \underline{q}_{g,n} &\leq q_{g,n} \leq \overline{q}_{g,n}\label{eq:ROPF-gen-const2}\\
     \qquad \underline{| v |} + \lambda\ &\leq\  |{\bf v}| \ \leq\  \overline{| v |} -\lambda\label{eq:ROPF-vol-const}\\
     \qquad &{\bf h}({\bf v}, {\bf p}, {\bf q}) = {\bf 0}\label{eq:ROPF-pfe}
\end{align}
\end{subequations}
Therefore, the generated solutions are guaranteed to be in the interior of the feasibility region with respect to the voltage magnitudes. Note that this data-driven approach approximates a mapping that deviates from the original AC OPF mapping. However, the training samples generated by solving~\eqref{eq:ROPF} are strictly in the interior of the voltage magnitudes feasibility set. Therefore, any bounded deviations in the learned mapping are expected to remain within the voltage limits. The parameter $\lambda$ can thus be considered as an algorithmic tuning parameter that addresses the optimality and feasibility trade-off. Notice that large values of $\lambda$ may render the original problem infeasible, and hence, the parameter $\lambda$ has to be tuned in order to obtain strictly feasible solutions.

In addition, to ensure the generation limits adhere to the prescribed AC OPF constraints, we use sigmoid functions at the output layer of the NN. This ensures that the output is bounded between $0$ and $1$ for the values of $\alpha_n$ and $\beta_n$. Designing the output layer of the neural network to be a sigmoid function acts as an implicit constraint on the learned AC OPF mapping, which illustrates the reasoning for reparameterizing the AC OPF solutions in terms of $\alpha_n$ and $\beta_n$. Although enforcing general constraints on the output of the neural network makes the training process challenging, incorporating a sigmoid function at the output layer does not increase the complexity of the training process.


\subsection{Generating training samples}
In a physical grid, historical AC OPF runs would be used as the training set for the neural network, yielding an abundance of data representing a wide variety of system states. However, for the purposes of simulation and testing, we must generate this set of training data. Let ${\bf p}_l \in \mathcal{R}^{|\mathcal{L}|}$ denote the vector collecting the load demand at all load buses. We generate the instances of load demand using the truncated Gaussian distribution 
\begin{align*}
    {\bf p}_l \sim \mathcal{T\!N}~\big({\bf p}_{l,0},\ \boldsymbol{\Sigma}_l,\ (1-\mu){\bf p}_{l,0},\ (1+\mu) {\bf p}_{l,0}\big),
\end{align*}
where ${\bf p}_{l,0}$ denotes a base loading profile, ${\boldsymbol{\Sigma}_l}$ denotes the covariance matrix, and $\mu$ denotes the maximum allowable deviation from the base load profile, which we set to $0.7$ in our simulations. The covariance matrix accounts for the relationships between loading patterns at different locations throughout the network. Here, we generated load profile realizations using a Gibbs sampler algorithm~\cite{kotecha1999gibbs}.
For considering deviations in the reactive power demand, we randomly chose the power factor of each load at each instance from a uniform distribution between $0.8$ and $1$.
Rarely, the generated load profile can result in an infeasible solution for the AC OPF problem, and hence infeasible for the R-AC OPF problem as well. In such cases, we discard this load profile from the training set, as it does not represent a physically realizable state of the power system.

\subsection{Learning model}
\begin{figure}
    \centering
    \includegraphics[width = 3.3in]{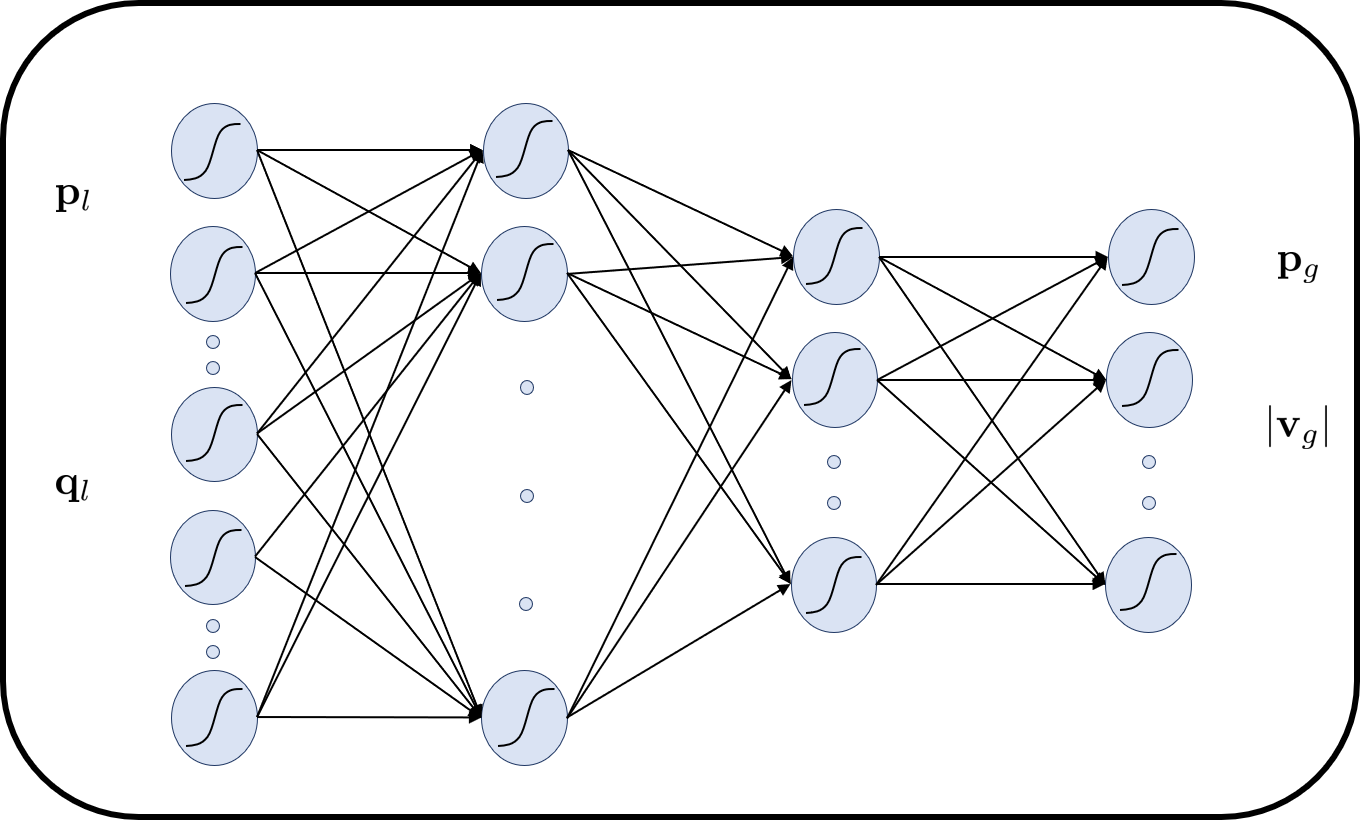}
    \caption{The proposed deep NN model.}
    \label{fig:NN_arch}
\end{figure}

A deep NN model (i.e., a NN with more than two layers) is used to approximate the mapping between the load profile and the optimal voltage magnitude and active power injections of the generators. The size of the input of the NN is $2|\mathcal{L}|$ representing the active and reactive load demand at the load buses, and the number of outputs is $2|\mathcal{G}|-1$ representing the voltage magnitude at the generator buses and the active power injection at all generators except the one at the slack/reference bus. Fig.~\ref{fig:NN_arch} depicts the architecture of the deep NN learning model utilized in our approach. Let $({\bf x}_i, {\bf y}_i)$ denote the $i$-th pair of data within the training set. The model is thus trained using the classical empirical loss minimization formulation which is given by
\begin{align*}
    \min_{\theta} \frac{1}{I}\sum_i^I \| {\bf y}_i - g({\bf x}_i; \theta) \|_2^2,
\end{align*}
where $I$ represents the total number of the training samples, $g(\cdot;\theta)$ represents the NN mapping where $\theta $ collects the trainable parameters.

\subsection{Recovering feasible solutions}
As mentioned in Section \ref{sec:ProbForm}, we pursue feasibility by both restricting the training set to include strictly feasible solutions and adhere to generation and voltage limits by constraining the output of the neural network. While this ensures feasibility of the learned values, the overall power flow equations and other generation constraints must be adhered to. Thus, the power flow equation are solved in order to recover and ensure feasibility of the overall AC OPF solution. Let $p^o_{g,n}$ and $|{v}^o_n|$ denote the output of the neural network representing the active power injection and voltage magnitude of bus $n\in \mathcal{G }$. Then, solving the power flow problem can be formulated as follows where the output of the NN representing the voltage magnitude and active power injection at generator $n$ are denoted by $|v_{n}^o|$ and $p_{g,n}^o$, respectively.
\begin{subequations}\label{eq:power-flow}
\begin{align}
    \text{find} & \qquad {\bf v}, {\bf p}_g, {\bf q}_g\\
    \text{subject to}& \qquad |v_n| = |{v}^o_n| \quad \forall n\in \mathcal{G}\\
    & \qquad p_{g,n} = p_{g, n}^o \quad \forall n\in \mathcal{G}\slash n_0\\
    & \qquad {\bf h}({\bf v}, {\bf p}_g, {\bf q}_g) = {\bf 0}
\end{align}
\end{subequations}

In rare cases, the solution of the power flow equations result in reactive power injections at the generators that exceed the limits in~\eqref{eq:gen-const2}. Let the solution of the power flow equations in~\eqref{eq:power-flow} representing the reactive power injection at bus $n\in \mathcal{G}$ be denoted by ${q}_{g,n}^o$. Also, denote the set of buses where the reactive power limits are violated by $\mathcal{G}_Q$. Then, we define $q_{g,n}^r$ at the buses $\mathcal{G}_Q$ as 
\begin{align}\label{eq:q_proj}
    q_{g,n}^r = \min(\overline{q}_{g,n}, ~\max(q_{g,n}^o, ~ \underline{q}_{g,n})).
\end{align}
Consequently, a modified power flow problem is to be solved in order to obtain the feasible operating point of the network which is formulated as
\begin{subequations}\label{eq:power-flow-ris}
\begin{align}
    \text{find} & \qquad {\bf v}, {\bf p}_g, {\bf q}_g\\
    \text{subject to}& \qquad |v_n| = |{v}^o_n| \quad \forall n\in \mathcal{G}\slash\mathcal{G}_Q\\
    & \qquad p_{g,n} = p_{g, n}^o \quad \forall n\in \mathcal{G}\slash n_0\\
    & \qquad q_{g, n} = q_{g, n}^r \quad \forall n \in \mathcal{G}_Q\\
    & \qquad {\bf h}({\bf v}, {\bf p}_g, {\bf q}_g) = {\bf 0}
\end{align}
\end{subequations}

The proposed approach is summarized in Algorithm ~\ref{alg:summary}. While an additional small amount of time is needed to solve the power flow equations and check the feasibility of all of the remaining variables, the overall solution procedure from end-to-end is \emph{extremely fast}, as we are dealing with a simpler problem which is solving the power equations and not solving a nonconvex optimization problem. In addition, note that the objective function value in the original problem \eqref{eq:OPF} is not altered by adjusting reactive power and voltage to preserve feasibility.

\SetKw{inputA}{Input: }
\SetKw{outputA}{Output: }
\DontPrintSemicolon
\begin{algorithm}[ht]
\setstretch{1.25}
	\caption{Proposed learning-based approach for AC OPF}
	{
		\inputA{ $ \{ p_{l,n}, q_{l,n} \}_{n\in\mathcal{L}} $}\;\\
		\outputA{$ \{ {\bf v}^r, {\bf p}_g^r, {\bf q}_g^r \} $}\;\\
		\Begin{
		$ \{|v_n^o|\}_{n\in\mathcal{G}}, \{p_{g,n}^o\}_{n\in\mathcal{G}\slash n_0} \longleftarrow {\bf g}( \{ p_{l,n}, q_{l,n} \}_{n\in\mathcal{L}}; \boldsymbol{\theta}) $ \;
		$ \{{\bf v}^o, {\bf p}_g^o, {\bf q}_q^o\} \longleftarrow \text{solution of \eqref{eq:power-flow}}  $\;\\
		\eIf{ \eqref{eq:gen-const2} is satisfied}{$\{ {\bf v}^r, {\bf p}_g^r, {\bf q}_g^r \} \longleftarrow \{{\bf v}^o, {\bf p}_g^o, {\bf q}_q^o\} $}
	    {
	    \For{$n \in \mathcal{G}_Q$}
	    {
	    $q_{g,n}^r \leftarrow \min\big\{\overline{q}_{g,n}, ~\max\{q_{g,n}^o, ~ \underline{q}_{g,n}\}\big\}$
	    }
	    $\{ {\bf v}^r, {\bf p}_g^r, {\bf q}_g^r \} \longleftarrow $ solution of~\eqref{eq:power-flow-ris}\;
		}
		}
	}
	\label{alg:summary}
\end{algorithm}

\section{Simulation Results}
In this section, we discuss the considered networks and analyze the optimality and feasibility obtained using the proposed approach. MATPOWER \cite{MATPOWER} was used to generate $100,000$ training samples for each considered network and to solve the power flow equations for the obtained optimal generator set-points.

\subsection{Test networks and setup}

Table \ref{table:Network-data} shows the number of nodes, generators, lines, and the base operating costs of the IEEE 118-, 57-, and 39-bus systems used in this paper. 
\begin{table}[htbp]
	\renewcommand{\arraystretch}{1.4}
	\caption[]{Details of the considered test networks.}
	\begin{center}
		\begin{tabular}{l  c c c c} 
			\toprule
			{\bf Test Case} &  {$|\mathcal{N}|$} & {$|\mathcal{G}|$} & {$|\mathcal{L}|$} & Base Cost (\$/hr) \\
			\midrule
			{\em IEEE 118-bus}    & $118$  & $54$ & $99$ & $129.66 \times 10^3$\\ 
			{\em IEEE 57-bus}     &  $57$ & $7$ & $42$ & $41.74 \times 10^3$\\ 
            {\em IEEE 39-bus}     & $39$  & $10$ & $21$ & $41.86 \times 10^3$ \\
			\bottomrule
		\end{tabular}
	\end{center}
	\label{table:Network-data}
\end{table}

For the NN network model, we utilized a NN with $3$ hidden layers with sigmoid activation function for all layers. The width of the first two hidden layers is equal to the number of inputs ($2 |\mathcal{L}|$) and the width of the third hidden layer is equal to the number of outputs ($2 |\mathcal{G}| -1$) for all networks. The learning model was implemented using the Python-based TensorFlow software library~\cite{tensorflow2015-whitepaper} and trained using the {Adam} optimizer~\cite{KingmaAdam}.

\subsection{Time comparisons}
We denote the time consumed to obtain an AC OPF solution using our method for the $t$-th test instance by $\tau_{{o},t}$, where this time includes the time needed to evaluate the neural network, the time needed to solve the power flow equations, and the time consumed by solving any additional power flow equations to ensure the feasibility of the solution. For comparison purposes, let $\tau_{\text{OPF}, t}$ denote the time consumed by the '{\sf PDIPM}' (primal-dual interior point method) solver to solve the original AC OPF problem optimally. In the considered experiments, this solver was on average the fastest and most successful solver. To measure the computational improvements, we define a speedup factor (SF) as
\begin{align*}
    \text{SF} = \frac{1}{T} \sum_{t=1}^{T} \frac{\tau_{\text{OPF},t}}{\tau_{o, t}},
\end{align*}
which we can use to measure the relative speedup of the learning-based framework versus solving the AC OPF directly. 

\begin{figure}
    \centering
    \includegraphics[width =3.2in]{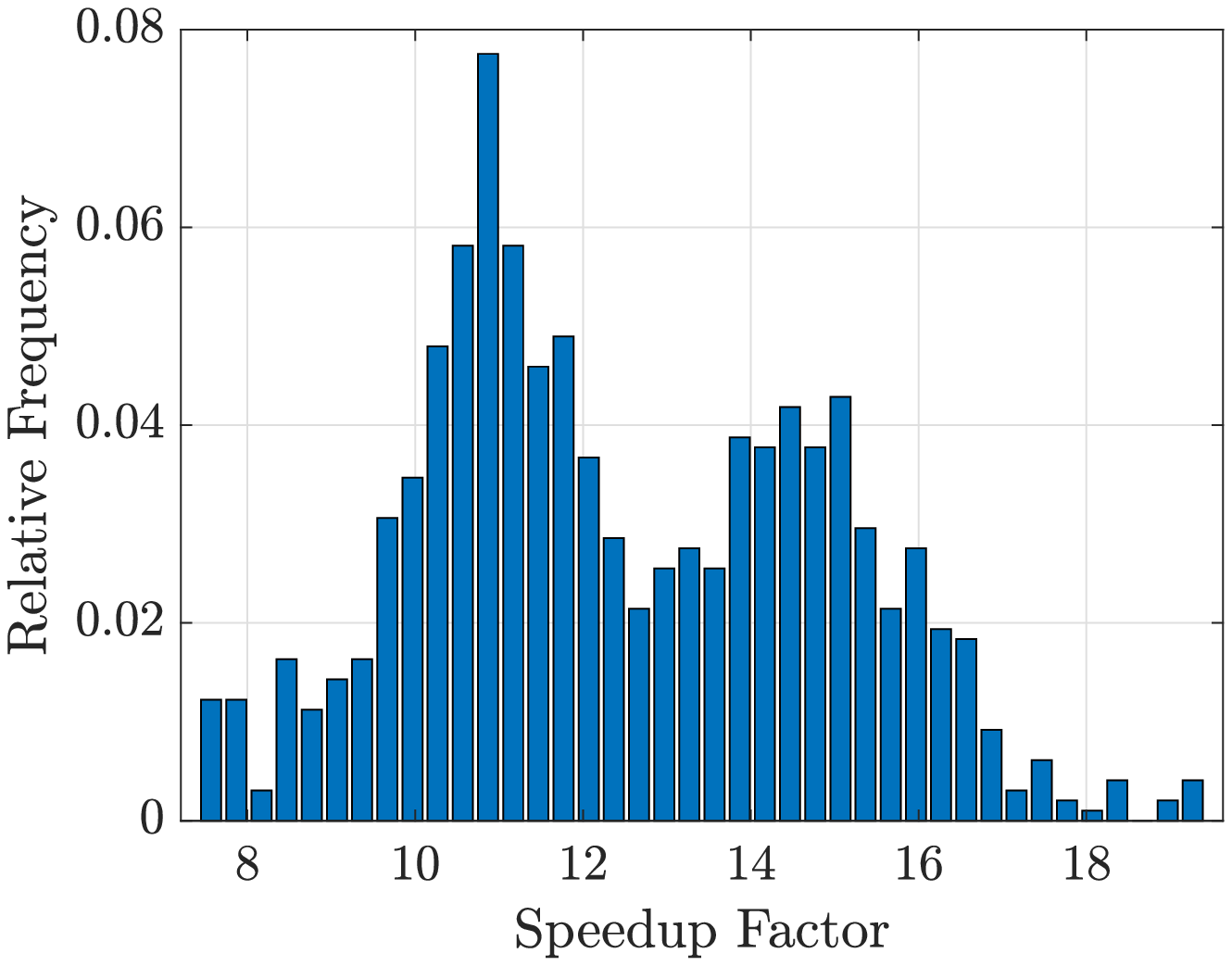}
    \caption{Relative histogram of speedup factor for solving AC OPF for IEEE 118-bus test case ($\lambda = 5 \times 10^{-3}$).}
    \label{fig:SF_hist}
\end{figure}

In Fig.~\ref{fig:SF_hist}, the relative histogram of the speedup factor is depicted for the IEEE 118-bus network with $\lambda = 0.005$. The speedup factor ranges from $6$x to $22$x for this case, which shows the potential of proposed approach to identify feasible AC OPF solutions extremely quickly, even in the slower cases. The two humps in the histogram result from the fact that the second stage of the algorithm is only needed for a portion of the test cases.   

\subsection{Feasibility and optimality}
After passing the network demands through the learning model, we obtain the generator set-points. Those set points are guaranteed to satisfy constraints~\eqref{eq:ROPF-gen-const} and~\eqref{eq:ROPF-vol-const} given the design of the output layer of the neural network. Further, to ensure feasibility of the \emph{overall} AC OPF solution, the infeasibility of the generated solution with respect to the reactive power limit constraints~\eqref{eq:ROPF-gen-const2} needs to be ensured by solving the power flow equations, as outlined in problem \eqref{eq:power-flow-ris}. We evaluate the infeasibility of this constraint for the solutions obtained directly using our learning model by defining a metric $\delta_q$ which is given by
\begin{align*}
    \delta_q = \frac{1}{T} \sum_{t=1}^{T} \frac{1}{|\mathcal{G}|} \| \boldsymbol{\xi}_{q,t} \|_2,
\end{align*}
where $\boldsymbol{\xi}_{q,t}$ collects all the reactive power limit violations across all buses. The $n$-th element of the vector $\boldsymbol{\xi}_{q,t}$ is given by
\begin{align*}
    \xi_{q,t,n} =  \max \{\underline{q}_{g,n} - q_{g,n}^o, 0\} + \max \{ q_{g,n}^o - \overline{q}_{g,n}, 0\}, 
\end{align*}
where $q_{g,n}^r$ denote the recovered reactive power injection at bus $n$ by solving the power flow equations at the set point given by the output of the neural network.

Let ${\bf p}_{g}^{\star}$ denote the optimal generation set-points obtained by solving the AC OPF problem~\eqref{eq:OPF}, and denote the generators' active power injection set-points obtained by our data-driven approach by ${\bf p}_g^{o}$. The average optimality of the solutions obtained through the proposed approach is evaluated with the following
\begin{align*}
\text{Optimality} = \frac{1}{T} \sum_{t=1}^{T} \frac{c ({\bf p}_{g,t}^{o}) - c ({\bf p}_{g,t}^{*})}{c ({\bf p}_{g,t}^{*}) },
\end{align*}
where $c ({\bf p}_{g,t}^{o})$ and $ c ({\bf p}_{g,t}^{\star})$ denote the total objective function value resulting from the solution obtained using our approach and the optimal cost found from solving the original AC OPF, respectively.

Denote by ${\bf p}_t^{o}$, ${\bf q}_t^{o}$, and ${\bf v}_t^o$ the solution obtained using our approach for a particular test scenario $t$. The feasibility of the solution is measured by assessing the satisfaction of the power flow equations. The maximum infeasibility is thus evaluated as follows:
\begin{align*}
    \text{Feasibility} = \max_{t} \| \bf{h}( {\bf v}_t^r, {\bf p}_t^r, {\bf q}_t^r ) \|_2.
\end{align*}

Tables \ref{table:Network-118}, \ref{table:Network-57}, and \ref{table:Network-39} illustrate the results from evaluating $1,000$ AC OPF test scenarios for each network. 
In the 118-bus test case, the initial infeasibility of the solution is higher when the value of lambda is set to be zero. This is attributed to the fact that the training samples in this case ($\lambda =0$) are on the boundary on the original problem feasibility set. For this reason, we did not include the results for ($\lambda =0$) for the other test cases.

As seen in all of the network simulations, the maximum infeasibility observed in the all of the test scenarios is very low (and well-within most solver tolerances for feasibility). The average computational speedup by using the data-driven approach versus solving the AC OPF directly is also significant, ranging from nearly a $8$x speedup factor up to over a $15$x SF. These tables also show the average infeasibility, $\delta_q$, of the reactive power limits \eqref{eq:ROPF-gen-const2} encountered, in MVAR, after solving the power flow equations with the learned generator active power set-points and voltages. Note, however, that the feasibility of these constraints is ensured using the procedure in Algorithm~\ref{alg:summary}, and the time required to do this is taken into account in the results.  

\begin{table}[t!]
	\renewcommand{\arraystretch}{1.4}
	\caption[]{Performance results for the IEEE 118-bus test case.}
	\begin{center}
		\begin{tabular}{l  c c c c} 
			\toprule
			{$\lambda$} &  {$\delta_q$} & {Optimality} & {SF} & Infeasibility \\
			\midrule
			{$ 0 $}    & $ 0.428 $  & $ 8.210 \times 10^{-5}$ & $ 7.97 $ & $ 2.63 \times 10^{-8}$\\ 
			{$ 0.005 $} & $ 0.302 $  & $2.974 \times 10^{-4} $ & $ 11.83 $ & $ 1.41 \times 10^{-8}$\\ 
            {$ 0.01 $} & $ 0.298 $  & $ 5.546 \times 10^{-4}$ & $11.74 $ & $3.58  \times 10^{-9}$\\ 
			\bottomrule
		\end{tabular}
	\end{center}
	\label{table:Network-118}
\end{table}

\begin{table}[t!]
	\renewcommand{\arraystretch}{1.4}
	\caption[]{Performance results for the IEEE 57-bus test case.}
	\begin{center}
		\begin{tabular}{l  c c c c} 
			\toprule
			{$\lambda$} &  {$\delta_q$} & {Optimality} & {SF} & Feasibility \\
			\midrule
		    {$ 0.005 $} & $ 1.58 $  & $4.568 \times 10^{-3} $ & $ 9.49  $ & $ 3.34 \times 10^{-8}$\\ 
            {$ 0.01 $} & $   1.37$  & $  7.049 \times 10^{-3}$ & $ 9.09 $ & $  3.75 \times 10^{-8}$\\ 
			\bottomrule
		\end{tabular}
	\end{center}
	\label{table:Network-57}
\end{table}

\begin{table}[t!]
	\renewcommand{\arraystretch}{1.4}
	\caption[]{Performance results for the IEEE 39-bus test case.}
	\begin{center}
		\begin{tabular}{l  c c c c} 
			\toprule
			{$\lambda$} &  {$\delta_q$} & {Optimality} & {SF} & Feasibility \\
			\midrule
		    {$ 0.005 $} & $ 1.32 $  & $3.259 \times 10^{-3} $ & $ 15.38 $ & $ 2.52 \times 10^{-8}$\\ 
            {$ 0.01 $} & $ 1.07 $  & $ 1.212 \times 10^{-2}$ & $12.86 $ & $2.78  \times 10^{-8}$\\ 
			\bottomrule
		\end{tabular}
	\end{center}
	\label{table:Network-39}
\end{table}

Further restricting the feasible set of voltages by increasing $\lambda$ from $0.005$ to $0.01$ generally has a positive impact on feasibility while mildly impacting the overall optimality of the solution. However, regardless of the $\lambda$ chosen here, the impact on the optimality of the learned solution is minimal for all considered networks, indicating that using the learning-assisted approach of recovering AC OPF solutions can preserve optimality and feasibility while providing significant computational speedups.

\section{Conclusions and Future Work}
In this paper, we presented a framework for solving AC OPF problems which utilizes machine learning to learn a mapping between system loading and optimal generation values. We provided a method to guarantee feasibility of the solution in the original AC OPF problem by training the model on strictly feasible solutions and then solving the power flow equations to recover the remainder of the complex network voltages and the reactive power generation values. From start to finish, the framework can recover a feasible solution to the AC OPF problem extremely quickly with a negligible sacrifice in optimality. Thus, the learning framework is overall faster and can provide a more accurate solution than the use of traditional methods that are used to solve AC OPF quickly such as linearizing or approximating the nonlinear power flow equations.

Future work includes exploring the efficacy of this framework on much larger networks (i.e., thousands of buses), and using learning to solve computationally intense, mixed integer problems such as security constrained AC OPF problems on faster timescales. The results in this paper also prove encouraging for using learning as a tool for quickly finding solutions to nonconvex optimization problems in areas outside of power systems. In potential applications where learning the optimal solution while ensuring feasibility is not possible or very difficult, we ask the reader to consider developing or using \emph{learning-assisted} or \emph{learning-enhanced} optimization techniques, wherein machine learning and optimization work hand-in-hand to find optimal solutions of the original problem, either by using learning to quickly find a starting point for the optimization, or by learning the solution of a subset of variables. 


\bibliographystyle{IEEEtran}
\bibliography{references.bib}

\begin{thebibliography}{10}
\providecommand{\url}[1]{#1}
\csname url@samestyle\endcsname
\providecommand{\newblock}{\relax}
\providecommand{\bibinfo}[2]{#2}
\providecommand{\BIBentrySTDinterwordspacing}{\spaceskip=0pt\relax}
\providecommand{\BIBentryALTinterwordstretchfactor}{4}
\providecommand{\BIBentryALTinterwordspacing}{\spaceskip=\fontdimen2\font plus
\BIBentryALTinterwordstretchfactor\fontdimen3\font minus
  \fontdimen4\font\relax}
\providecommand{\BIBforeignlanguage}[2]{{%
\expandafter\ifx\csname l@#1\endcsname\relax
\typeout{** WARNING: IEEEtran.bst: No hyphenation pattern has been}%
\typeout{** loaded for the language `#1'. Using the pattern for}%
\typeout{** the default language instead.}%
\else
\language=\csname l@#1\endcsname
\fi
#2}}
\providecommand{\BIBdecl}{\relax}
\BIBdecl

\bibitem{Molzahn_list}
D.~K. {Molzahn}, F.~{D\"{o}rfler}, H.~{Sandberg}, S.~H. {Low},
  S.~{Chakrabarti}, R.~{Baldick}, and J.~{Lavaei}, ``A survey of distributed
  optimization and control algorithms for electric power systems,'' \emph{IEEE
  Transactions on Smart Grid}, vol.~8, no.~6, pp. 2941--2962, Nov 2017.

\bibitem{Lavaei10}
J.~{Lavaei} and S.~H. {Low}, ``Convexification of optimal power flow problem,''
  in \emph{2010 48th Annual Allerton Conference on Communication, Control, and
  Computing (Allerton)}, Sep. 2010, pp. 223--232.

\bibitem{swaroop2015linear}
S.~Guggilam, E.~Dall'Anese, Y.~Chen, S.~Dhople, and G.~B. Giannakis, ``Scalable
  optimization methods for distribution networks with high {PV} integration,''
  \emph{{IEEE} Transactions on Smart Grid}, vol.~7, no.~4, July 2016.

\bibitem{bolognani2015linear}
S.~Bolognani and F.~D\"orfler, ``Fast power system analysis via implicit
  linearization of the power flow manifold,'' in \emph{2015 53rd Annual
  Allerton Conf. on Communication, Control, and Computing}, Oct. 2015.

\bibitem{linModels}
A.~Bernstein and E.~Dall'Anese, ``Linear power-flow models in multiphase
  distribution networks,'' in \emph{{The 7th IEEE Intl. Conf. on Innovative
  Smart Grid Technologies}}, Sep. 2017.

\bibitem{Erseghe14}
T.~{Erseghe}, ``Distributed optimal power flow using {ADMM},'' \emph{IEEE
  Transactions on Power Systems}, vol.~29, no.~5, pp. 2370--2380, Sep. 2014.

\bibitem{Zheng_Dist_16}
W.~{Zheng}, W.~{Wu}, B.~{Zhang}, H.~{Sun}, and Y.~{Liu}, ``A fully distributed
  reactive power optimization and control method for active distribution
  networks,'' \emph{IEEE Transactions on Smart Grid}, vol.~7, no.~2, pp.
  1021--1033, March 2016.

\bibitem{BakerMPC}
K.~Baker, J.~Guo, G.~Hug, and X.~Li, ``Distributed {MPC} for efficient
  coordination of storage and renewable energy sources across control areas,''
  \emph{IEEE Transactions on Smart Grid}, vol.~7, no.~2, pp. 992--1001, Mar.
  2016.

\bibitem{Tang17}
Y.~{Tang}, K.~{Dvijotham}, and S.~{Low}, ``Real-time optimal power flow,''
  \emph{IEEE Transactions on Smart Grid}, vol.~8, no.~6, pp. 2963--2973, Nov
  2017.

\bibitem{opfPursuit}
E.~Dall'Anese and A.~Simonetto, ``Optimal power flow pursuit,'' \emph{{IEEE}
  {T}ransactions on {S}mart {G}rid}, vol.~9, no.~2, pp. 942--952, March 2018.

\bibitem{Hauswirth17}
A.~{Hauswirth}, A.~{Zanardi}, S.~{Bolognani}, F.~{D\"{o}rfler}, and G.~{Hug},
  ``Online optimization in closed loop on the power flow manifold,'' in
  \emph{2017 IEEE Manchester PowerTech}, June 2017.

\bibitem{Bernstein_online_19}
A.~Bernstein and E.~Dall'Anese, ``Real-time feedback-based optimization of
  distribution grids: A unified approach,'' 2019, [Online] Available:
  https://arxiv.org/abs/1711.01627.

\bibitem{Baker_jointCC}
K.~Baker and A.~Bernstein, ``Joint chance constraints in {AC} optimal power
  flow: Improving bounds through learning,'' \emph{IEEE Transactions on Smart
  Grid (to appear)}, 2019.

\bibitem{gunduz2019machine}
D.~Gunduz, P.~de~Kerret, N.~D. Sidiropoulos, D.~Gesbert, C.~Murthy, and
  M.~van~der Schaar, ``Machine learning in the air,'' \emph{arXiv preprint
  arXiv:1904.12385}, 2019.

\bibitem{wang2018review}
Y.~Wang, Q.~Chen, T.~Hong, and C.~Kang, ``Review of smart meter data analytics:
  Applications, methodologies, and challenges,'' \emph{IEEE Trans. Smart Grid},
  vol.~10, no.~3, pp. 3125--3148, 2018.

\bibitem{Learning_ACOPF}
N.~Guha, Z.~Wang, and A.~Majumdar, ``Machine learning for {AC} optimal power
  flow,'' in \emph{Climate Change Workshop at The Thirty-sixth International
  Conference on Machine Learning (ICML)}, June 2019.

\bibitem{Zamzam-19physics}
A.~S. Zamzam and N.~D. Sidiropoulos, ``Physics-aware neural networks for
  distribution system state estimation,'' 2019, [Online] Available:
  https://arxiv.org/abs/1903.09669.

\bibitem{yang2019real}
Q.~Yang, G.~Wang, A.~Sadeghi, G.~B. Giannakis, and J.~Sun, ``Real-time voltage
  control using deep reinforcement learning,'' 2019, [Online] Available:
  https://arxiv.org/abs/1904.09374.

\bibitem{Data18}
H.~Akhavan-Hejazi and H.~Mohsenian-Rad, ``Power systems big data analytics: An
  assessment of paradigm shift barriers and prospects,'' \emph{Energy Reports},
  vol.~4, pp. 91 -- 100, 2018.

\bibitem{Zamzam19}
A.~S. {Zamzam}, X.~{Fu}, and N.~D. {Sidiropoulos}, ``Data-driven learning-based
  optimization for distribution system state estimation,'' \emph{IEEE
  Transactions on Power Systems (to appear)}, 2019.

\bibitem{Baker_learning}
K.~Baker, ``Learning warm-start points for {AC} optimal power flow,'' in
  \emph{IEEE Machine Learning for Signal Processing (MLSP) Conference},
  Pittsburgh, PA 2019.

\bibitem{Misra19}
S.~Misra, L.~Roald, and Y.~Ng, ``Learning for constrained optimization:
  Identifying optimal active constraint sets,'' 2019, [Online] Available at:
  https://arxiv.org/pdf/1802.09639.

\bibitem{Deka19}
D.~Deka and S.~Misra, ``Learning for {DC-OPF}: Classifying active sets using
  neural nets,'' 2019, [Online] Available at: https://arxiv.org/pdf/1902.05607.

\bibitem{Bertsimas_learn_19}
D.~Bertsimas and B.~Stellato, ``Online mixed-integer optimization in
  milliseconds,'' July 2019, [Online] Available at:
  https://arxiv.org/pdf/1907.02206.

\bibitem{DeepOPF}
X.~Pan, T.~Zhao, and M.~Chen, ``Deep{OPF}: Deep neural network for {DC} optimal
  power flow,'' August 2019, [Online] Available at:
  https://arxiv.org/pdf/1905.04479.

\bibitem{Araposthatis-1981}
A.~Araposthatis, S.~Sastry, and P.~Varaiya, ``Analysis of power-flow
  equation,'' \emph{International Journal of Electrical Power \& Energy
  Systems}, vol.~3, no.~3, pp. 115--126, July 1981.

\bibitem{zhou2019optimal}
F.~Zhou, J.~Anderson, and S.~H. Low, ``The optimal power flow operator: Theory
  and computation,'' \emph{arXiv preprint arXiv:1907.02219}, 2019.

\bibitem{grohs2019deep}
P.~Grohs, D.~Perekrestenko, D.~Elbr{\"a}chter, and H.~B{\"o}lcskei, ``Deep
  neural network approximation theory,'' \emph{arXiv preprint
  arXiv:1901.02220}, 2019.

\bibitem{molzahn2015semidefinite}
D.~K. Molzahn, S.~S. Baghsorkhi, and I.~A. Hiskens, ``Semidefinite relaxations
  of equivalent optimal power flow problems: An illustrative example,'' in
  \emph{IEEE International Symposium on Circuits and Systems (ISCAS)}, May
  2015, pp. 1887--1890.

\bibitem{kotecha1999gibbs}
J.~H. Kotecha and P.~M. Djuric, ``Gibbs sampling approach for generation of
  truncated multivariate gaussian random variables,'' in \emph{1999 IEEE
  International Conference on Acoustics, Speech, and Signal Processing},
  vol.~3, 1999, pp. 1757--1760.

\bibitem{MATPOWER}
R.~D. Zimmerman, C.~E. Murillo-Sanchez, and R.~J. Thomas, ``{MATPOWER}:
  Steady-state operations, planning, and analysis tools for power systems
  research and education,'' \emph{IEEE Trans. on Power Systems}, vol.~26,
  no.~1, pp. 12--19, Feb 2011.

\bibitem{tensorflow2015-whitepaper}
\BIBentryALTinterwordspacing
M.~Abadi \emph{et~al.}, ``{TensorFlow}: Large-scale machine learning on
  heterogeneous systems,'' 2015, software available from tensorflow.org.
  [Online]. Available: \url{https://www.tensorflow.org/}
\BIBentrySTDinterwordspacing

\bibitem{KingmaAdam}
D.~P. Kingma and J.~Ba, ``Adam: {A} method for stochastic optimization,'' in
  \emph{3rd International Conference on Learning Representations, {ICLR} 2015,
  San Diego, CA, USA, May}, 2015.

\end{thebibliography}

\end{document}